\documentclass[letterpaper, 10 pt, conference]{ieeeconf}  

\IEEEoverridecommandlockouts                              

\usepackage{graphics} 
\usepackage{amsmath} 
\usepackage{amssymb}  

\usepackage[nolist]{acronym}
\usepackage{tabularx}
\usepackage{multirow}
\usepackage{units}
\usepackage{pgfplots}
\graphicspath{{illustrations/}}
\usepackage{url}

\newcommand\MYhyperrefoptions{bookmarks=true,bookmarksnumbered=true,
	pdfpagemode={UseOutlines},plainpages=false,pdfpagelabels=true,
	colorlinks=true,linkcolor={black},citecolor={black},urlcolor={black},
	pdftitle={Fusing Visual Appearance and Geometry for Multi-modality 6DoF Object Tracking}}
\makeatletter
\let\NAT@parse\undefined
\makeatother
\usepackage[\MYhyperrefoptions,pdftex]{hyperref}

\title{\LARGE \bf
Fusing Visual Appearance and Geometry for\\ Multi-modality 6DoF Object Tracking
}

\author{Manuel Stoiber$^{1, 3}$, Mariam Elsayed$^{3}$, Anne E. Reichert$^{1, 3}$, Florian Steidle$^{1, 3}$,\\ Dongheui Lee$^{2}$, and Rudolph Triebel$^{1, 3}$
	\thanks{$^{1}$ Institute of Robotics and Mechatronics, German Aerospace Center, 82234 Wessling, Germany, {\tt\small \{firstname.lastname\}@dlr.de}}%
	\thanks{$^{2}$ Autonomous Systems Group, Vienna University of Technology, 1040 Vienna, Austria, {\tt\small dongheui.lee@tuwien.ac.at}}
	\thanks{$^{3}$ Technical University of Munich, 80333 Munich, Germany}%
}

\begin{acronym}
	\acro{ICP}[ICP]{Iterative Closest Point}
	\acro{PDF}[PDF]{probability density function}
	\acro{6DoF}[6DoF]{six degrees of freedom}
	\acro{CNN}[CNN]{convolutional neural network}
	\acro{EKF}[EKF]{Extended Kalman Filter}
	\acro{SfM}[SfM]{Structure from Motion}
	\acro{SLAM}[SLAM]{Simultaneous Localization And Mapping}
	\acro{UKF}[UKF]{Unscented Kalman Filter}
	\acro{YCB}[YCB]{Yale-CMU-Berkeley}
\end{acronym}

\begin{document}

\maketitle
\thispagestyle{empty}
\pagestyle{empty}


\begin{abstract}
	In many applications of advanced robotic manipulation, \ac{6DoF} object pose estimates are continuously required.
	In this work, we develop a multi-modality tracker that fuses information from visual appearance and geometry to estimate object poses.
	The algorithm extends our previous method \textit{ICG}, which uses geometry, to additionally consider surface appearance.
	In general, object surfaces contain local characteristics from text, graphics, and patterns, as well as global differences from distinct materials and colors.
	To incorporate this visual information, two modalities are developed.
	For local characteristics, keypoint features are used to minimize distances between points from keyframes and the current image.
	For global differences, a novel region approach is developed that considers multiple regions on the object surface.
	In addition, it allows the modeling of external geometries.
	Experiments on the \textit{YCB-Video} and \textit{OPT} datasets demonstrate that our approach \textit{ICG+} performs best on both datasets, outperforming both conventional and deep learning-based methods.
	At the same time, the algorithm is highly efficient and runs at more than 300\,Hz.
	The source code of our tracker is publicly available.
\end{abstract}


\section{Introduction}\label{sec:in}
\PARstart{T}{racking} rigid objects and estimating their \ac{6DoF} pose is a fundamental problem in computer vision that is highly important for many areas of
robotics.
Applications range from advanced manipulation tasks with unpredictable object motion to novel robotic designs without perfect forward kinematics.
In many applications, \ac{6DoF} object tracking is thereby used together with visual servoing techniques to close the perception-action loop and facilitate new capabilities.
In contrast to global detection and pose estimation, the goal is to continuously estimate the rotation and translation of an object relative to the camera from consecutive images and an object model.
Given various challenges, such as occlusions, cluttered environments, object symmetries, and limited computational resources, many techniques have been developed \cite{Lepetit2005, Yilmaz2006}.
In general, they can be categorized by their use of edges, keypoints,
direct optimization, object regions, depth information, and deep learning.

Edge-based techniques fit edges from the object model to high-intensity gradients in the image \cite{Harris1990, Bugaev2018}.
While such methods were very popular in the past, they often struggle with texture and background clutter.
Keypoint-based methods, on the other hand, use characteristic points on the object’s surface \cite{Vacchetti2004, Lourakis2013}.
Given strong texture, they typically provide a large basin of convergence and are robust to illumination changes.
Methods based on direct optimization \cite{Crivellaro2014, Zhong2019} minimize a pixel-wise error over the object silhouette.
While they often have a smaller basin of convergence and are less robust to illumination changes, they perform better in cases of weak texture.
For the tracking of textureless objects in cluttered environments, region-based methods \cite{Prisacariu2012, Stoiber2021} provide very good results.
They fit an object model to best explain the segmentation between the object and the background.
In contrast to edge-based techniques, existing methods only consider a single object region and neglect information inside the contour.
With depth cameras, algorithms that minimize the distance between an object model and depth measurement have also been developed \cite{Schmidt2015a, Krainin2011}.
Given high-quality images, they typically perform very well.
Recently, techniques based on deep learning \cite{Wen2020, Deng2021} have also been proposed.
While they hold great promise for the future, they require significant computational resources and large amounts of training data.
Also, it is unclear how well they perform compared to conventional techniques.

\begin{figure*}[t]
	\vspace{1ex}
	\centering
	\input{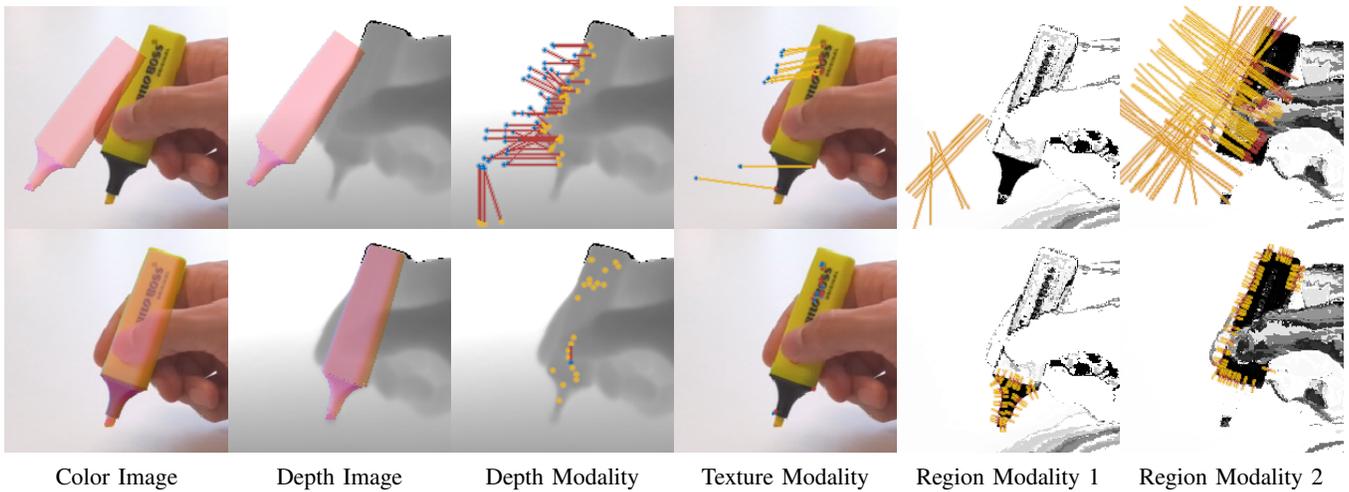}
	\vspace{-1.8ex}
	\caption{
		Combination of multiple modalities for the tracking of a marker pen.
		In the first row, the initial state of the optimization is shown, while the second row visualizes the final state.
		For the color and depth image, a rendered overlay illustrates the object in the predicted pose.
		In the visualization of the depth modality, correspondences between blue model points and yellow depth image points are illustrated by red lines.
		For the texture modality, keyframe features are given in blue, while detected points from the current frame are shown in red.
		Matched features are connected by yellow lines.
		For the two region modalities, which consider the pen's yellow body and black tip, probabilities that pixels belong to the respective object region are encoded in grayscale images.
		Pixels considered by the method's so-called correspondence lines are illustrated in yellow, with high probabilities for the location of the contour indicated in red.
		While longer lines with a larger scale are used in the first iteration, smaller lines are used in the final iteration.
	}\label{fig:in00}
\end{figure*}

In various publications, it was shown that a combination of different techniques is highly beneficial, both for robustness and tracking quality \cite{Brox2010, Krainin2011, Zhong2019}.
Instead of developing an independent technique, we therefore extend our previous approach \textit{Iterative Corresponding Geometry (ICG)} \cite{Stoiber2022}, which incorporates depth and single-region information.
While \textit{ICG} performs very well for a wide range of objects, it only uses geometry without considering visual information on the object's surface.
As a consequence, if the visible geometry in the vicinity of a particular pose is not conclusive, \textit{ICG} is not able to predict the correct result.
To overcome this limitation, we propose the incorporation of an additional texture modality that uses keypoint features.
In addition to this new modality, we also extend the existing region approach to multi-region tracking.
This allows our method to utilize information inside the object's contour.
While both improvements help to overcome ambiguities in the object's geometry, they utilize different sources of visual information and are thus able to complement each other.
An illustration of the resulting multi-modality tracker, which we call \textit{ICG+}, is shown in Fig.~\ref{fig:in00}.
The approach allows to combine depth, texture, and multi-region information from multiple cameras in a flexible framework.
In experiments on the \textit{YCB-Video} \cite{Xiang2018} and \textit{OPT} \cite{Wu2017} datasets, we demonstrate that the developed algorithm significantly improves the current state of the art and outperforms all existing methods.
At the same time, it is highly efficient and runs at more than 300\,Hz.


\section{Related Work}\label{sec:r}

In the following section, we provide an overview of region- and keypoint-based techniques and introduce modern deep learning-based approaches.
Region-based methods use image statistics to differentiate between a foreground that corresponds to the object silhouette and a background.
One of the most prominent methods based on this technique is \textit{PWP3D} \cite{Prisacariu2012}.
It inspired many subsequent approaches that proposed various modifications.
For example, to better differentiate between foreground and background, methods that localize the statistical modeling were developed \cite{Hexner2016, Tjaden2018, Zhong2020}.
Also, contour constraints were suggested that explicitly deal with partial occlusions and color ambiguities \cite{Huang2021}.
With respect to optimization, different techniques, including Levenberg-Marquardt \cite{Prisacariu2015}, Gauss-Newton \cite{Tjaden2018}, and Newton with Tikhonov regularization \cite{Stoiber2020b}, were considered.
The relatively poor efficiency of region-based methods was addressed with the development of \textit{SRT3D} \cite{Stoiber2020b, Stoiber2021}.
It features a highly-efficient sparse formulation.
For evaluation purposes, the \textit{OPT} \cite{Wu2017}, \textit{RBOT} \cite{Tjaden2018}, and \textit{BCOT} \cite{Li2022} datasets are often used.
Finally, combined approaches were also proposed.
For this, methods that incorporate depth data proved particularly successful \cite{Kehl2017, Ren2017, Stoiber2022}.
Also, edge-based algorithms were used \cite{Li2021, Sun2021}.
To consider texture, direct optimization techniques that use pixel colors \cite{Liu2020} or descriptor fields \cite{Zhong2019} were also developed.

In our work, we use keypoint features to incorporate texture information.
Keypoint features mark distinct locations on the object surface and provide a descriptor that depends on the surrounding.
For object tracking, most methods reconstruct 3D points and establish 2D-3D correspondences.
They then optimize for the pose that either minimizes the reprojection error in the image \cite{Vacchetti2004, Lourakis2013} or an error in 3D space \cite{Brox2010}.
Required 3D points are typically reconstructed from a 3D mesh model \cite{Brox2010} or from pre-registered reference frames \cite{Vacchetti2004, Lourakis2013}.
Approaches that minimize the 3D point-to-point distance \cite{Krainin2011} or that adopt local bundle adjustment \cite{Vacchetti2004} were also proposed.
For keypoint features, a wide variety of detectors and descriptors has been developed.
Early work can thereby be traced back to Moravec and Harris.
Especially with the development of \textit{SIFT} \cite{Lowe2004}, which is both highly accurate and robust, the use of keypoint features became very popular.
Approaches similar to \textit{SIFT} that improve computational efficiency were later presented with  \textit{SURF} \cite{Bay2006} or \textit{DAISY} \cite{Tola2010}.
Also, to improve the speed of feature detection, the intensity-based algorithm \textit{FAST} \cite{Rosten2005} was developed.
It was further improved with the \textit{ORB} \cite{Rublee2011} detector, which selects features using the Harris response.
Together with lightweight binary descriptors such as \textit{BRIEF} \cite{Calonder2010}, \textit{ORB} \cite{Rublee2011}, \textit{BRISK} \cite{Leutenegger2011}, or \textit{FREAK} \cite{Alahi2012}, highly efficient algorithms can be designed.
In recent years, variants that use deep learning at various stages have also been developed.
Prominent examples are \textit{SuperPoint} \cite{DeTone2018}, \textit{D2-Net} \cite{Dusmanu2019}, or \textit{LoFTR} \cite{Sun2021b}.
However, while those techniques show excellent results, they are often less efficient than many conventional methods and are therefore not considered for our approach.

Finally, deep learning has also been used to directly infer the object pose.
Most prominent are thereby approaches that predict the relative pose between an object rendering and the camera image \cite{Manhardt2018,Li2018,Wen2020}.
In addition to those render-and-compare algorithms, \textit{PoseRBPF} \cite{Deng2021} uses a Rao-Blackwellized particle filter and pose-representative latent codes \cite{Sundermeyer2018}.
Also, \textit{TP-AE} \cite{Zheng2022} proposed a temporally primed framework with auto encoders, while \textit{ROFT} \cite{Piga2022} synchronizes low framerate pose estimates with fast optical flow.


\section{Approach}\label{sec:c}
In this section, we first present basic mathematical concepts.
This is followed by an introduction to our multi-modality tracking approach.
Subsequently, mathematics for the keypoint-based texture modality are derived.
Finally, after a short introduction to \textit{ICG}'s region modality, our extension to multi-region tracking is described.

\subsection{Preliminaries}\label{sec:c0}
In this work, $\pmb{X} = \begin{bmatrix} X& Y& Z\end{bmatrix}^\top\in \mathbb{R}^3$ denotes a 3D model point, with the corresponding homogeneous vector defined as $\pmb{\widetilde{X}} = \begin{bmatrix} X& Y& Z& 1\end{bmatrix}^\top$.
In image space, 2D coordinates $\pmb{x} = \begin{bmatrix} x& y\end{bmatrix}^\top \in \mathbb{R}^2$ are used to describe pixel locations.
The projection of a 3D model point $\pmb{X}$ into an undistorted image is described by the pinhole camera model
\begin{equation}\label{eq:c01}
	\renewcommand\arraystretch{1.2}
	\pmb{x} = \pmb{\pi}(\pmb{X}) = 
	\begin{bmatrix}
		\frac{X}{Z}f_x + p_x\\
		\frac{Y}{Z}f_y + p_y
	\end{bmatrix},
\end{equation}
where $f_x$ and $f_y$ are the focal lengths and $p_x$ and $p_y$ the principal point coordinates.

The relative pose between the model and camera coordinate frames $\textrm{M}$ and $\textrm{C}$ can be described using the homogeneous transformation matrix ${}_C\pmb{T}_M{}\in \mathbb{SE}(3)$ as follows
\begin{equation}\label{eq:c03}
	_\textrm{C}\pmb{\widetilde{X}} = {}_\textrm{C}\pmb{T}_\textrm{M}\,{}_\textrm{M}\pmb{\widetilde{X}} =
	\begin{bmatrix}
		_\textrm{C}\pmb{R}_\textrm{M} & _\textrm{C}\pmb{t}_\textrm{M} \\ \pmb{0} & 1
	\end{bmatrix}
	{}_\textrm{M}\pmb{\widetilde{X}},
\end{equation}
where ${}_\textrm{M}\pmb{\widetilde{X}}$ and $_\textrm{C}\pmb{\widetilde{X}}$ are 3D points in the model and camera coordinate frames.
Also, $_\textrm{C}\pmb{R}_\textrm{M} \in \mathbb{SO}(3)$ and $_\textrm{C}\pmb{t}_\textrm{M} \in \mathbb{R}^3$ are the rotation matrix and the translation vector that define the transformation from the coordinate frame $\textrm{M}$ to $\textrm{C}$.

For small variations of the pose in the model reference frame $\textrm{M}$, we use the following minimal parameterization derived from the axis-angle representation
\begin{equation}\label{eq:c04}
	_\textrm{M}\pmb{\widetilde{X}}(\pmb{\theta}) =
	{}_\textrm{M}\pmb{T}(\pmb{\theta})\,{}_\textrm{M}\pmb{\widetilde{X}} =
	\begin{bmatrix}
		\pmb{I} + [\pmb{\theta}_\textrm{r}]_\times & \pmb{\theta}_\textrm{t} \\ \pmb{0} & 1
	\end{bmatrix}
	{}_\textrm{M}\pmb{\widetilde{X}},
\end{equation}
where $[\pmb{\theta}_\textrm{r}]_\times$ is the cross-product matrix of $\pmb{\theta}_\textrm{r}$, and the rotation vector $\pmb{\theta}_\textrm{r}\in \mathbb{R}^3$ and the translation vector $\pmb{\theta}_\textrm{t}\in \mathbb{R}^3$ are components of the variation vector $\pmb{\theta}^\top = \begin{bmatrix} \pmb{\theta}_\textrm{r}^\top & \pmb{\theta}_\textrm{t}^\top \end{bmatrix}$.
Note that the variation in the model frame $\textrm{M}$ ensures that multiple cameras with different camera frames $\textrm{C}$ can be used.

\subsection{Multi-modality Tracking}\label{sec:c1}
In our approach, we use a \ac{PDF} to express the probability of a particular pose given data from multiple cameras.
The pose is then estimated by maximizing this function.
For our multi-modality tracker \textit{ICG+}, the joint \ac{PDF} over all modalities is defined as follows,
\begin{equation}\label{eq:c10}
	p(\pmb{\theta} \mid \pmb{\mathcal{D}}) = \prod_{i=1}^{n_\textrm{d}} p(\pmb{\theta} \mid \pmb{\mathcal{D}}_{\textrm{d}i}) \prod_{i=1}^{n_\textrm{t}} p(\pmb{\theta} \mid \pmb{\mathcal{D}}_{\textrm{t}i}) \prod_{i=1}^{n_\textrm{r}} p(\pmb{\theta} \mid \pmb{\mathcal{D}}_{\textrm{r}i}),
\end{equation}
where $\pmb{\mathcal{D}}_{\textrm{d}i}$ considers data from a depth modality, $\pmb{\mathcal{D}}_{\textrm{t}i}$ is data from a texture modality, and $\pmb{\mathcal{D}}_{\textrm{r}i}$ denotes data from a region modality.
Note that while for texture and depth, only one modality per camera is used, for the derived multi-region approach, multiple modalities can be configured.

To maximize this \ac{PDF} and estimate the pose variation, Newton optimization with Tikhonov regularization is used
\begin{equation}\label{eq:c11}
	\pmb{\hat{\theta}} = \bigg(-\pmb{H} + 
	\begin{bmatrix}
		\lambda_\textrm{r} \pmb{I}_3 & \pmb{0}\\
		\pmb{0} & \lambda_\textrm{t} \pmb{I}_3
	\end{bmatrix}
	\bigg)^{-1}\pmb{g},
\end{equation}
where $\lambda_\textrm{r}$ and $\lambda_\textrm{t}$ are rotational and translational regularization parameters.
Also, the gradient $\pmb{g}$ and the Hessian $\pmb{H}$ are the first- and second-order derivatives of the logarithm of the joint \ac{PDF} $\ln\big(p(\pmb{\theta}\mid\pmb{\mathcal{D}})\big)$ evaluated at $\pmb{\theta} = \pmb{0}$.
Based on individual modalities, they can be assembled as follows
\begin{align}\label{eq:c12}
	\pmb{g} &= \sum_{i=1}^{n_\textrm{d}} \pmb{g}_{\textrm{d}i} + \sum_{i=1}^{n_\textrm{t}} \pmb{g}_{\textrm{t}i} + \sum_{i=1}^{n_\textrm{r}} \pmb{g}_{\textrm{r}i},\\[5pt]\label{eq:mmt:c13}
	\pmb{H} &= \sum_{i=1}^{n_\textrm{d}} \pmb{H}_{\textrm{d}i} + \sum_{i=1}^{n_\textrm{t}} \pmb{H}_{\textrm{t}i} + \sum_{i=1}^{n_\textrm{r}} \pmb{H}_{\textrm{r}i}.
\end{align}
The gradient and Hessian of the texture modality will be derived in the subsequent section.
For depth and region modalities, we use the same formulation as \textit{ICG} and refer interested readers to the corresponding publication \cite{Stoiber2022}.

Finally, given the calculated variation vector $\pmb{\hat{\theta}}$, pose estimates can be updated as follows
\begin{equation}\label{eq:c14}
	_\textrm{C}\pmb{T}_\textrm{M}^+ =
	{}_\textrm{C}\pmb{T}_\textrm{M}
	\begin{bmatrix}
		\exp([\pmb{\hat{\theta}}_\textrm{r}]_\times) & \pmb{\hat{\theta}}_\textrm{t} \\ \pmb{0} & 1
	\end{bmatrix}.
\end{equation}
By iteratively repeating this process, one is able to find the pose that maximizes the joint \ac{PDF}, considering depth, texture, and multi-region information from multiple cameras.

\subsection{Texture Modality}\label{sec:c2}
To consider local object appearance, we design a \ac{PDF} that uses keypoint features.
For our modality, points are detected on each new frame within a rectangular region close to the previous pose estimate.
Descriptors from those points are then matched to features from keyframes.
A frame is considered a keyframe if the orientational difference to existing keyframes exceeds a certain threshold.
Note that the use of keyframes helps to reduce drift and makes the overall tracking more robust.
Finally, if a frame is considered a keyframe, a depth rendering is generated, and for each keypoint that falls on the silhouette, 3D model points are reconstructed.
Together with their descriptors, unoccluded 3D points are then stored for the keyframe.

Given detected 2D points $\pmb{x}'_i$ from the current frame and matching 3D model points ${}_\textrm{M}\pmb{X}_i$ from keyframes, a \ac{PDF} can be formulated.
Similar to other methods \cite{Vacchetti2004}, we assume a normal distribution for the reprojection error and write the following function for $n$ independent point pairs
\begin{equation}\label{eq:c22}
	p(\pmb{\theta} \mid \pmb{\mathcal{D}}_\textrm{t}) \propto \prod_{i=1}^{n} \exp\bigg(-\frac{1}{2{\sigma_\textrm{t}}^2}\rho_\textrm{tuk}\Big(\big\lVert\pmb{x}'_i - \pmb{x}_i(\pmb{\theta})\big\rVert_2\Big)\bigg),
\end{equation}
with 3D model points projected into the image space
\begin{equation}\label{eq:c23}
\pmb{x}_i(\pmb{\theta}) = \pmb{\pi}\big({}_\textrm{C}\pmb{T}_\textrm{M}\, {}_\textrm{M}\pmb{T}(\pmb{\theta})\,{}_\textrm{M}\pmb{X}_i\big).
\end{equation}

The user-defined standard deviation $\sigma_\textrm{t}$ takes into account the uncertainty of the texture modality compared to other modalities.
Also, the term $\rho_\textrm{tuk}$ denotes the Tukey norm.
It minimizes the effect of outliers and is defined as
\begin{equation}\label{eq:c24}
	\rho_\textrm{tuk}(x) =
	\begin{cases}
		\frac{c^2}{6}\big(1 - (1 - (\frac{x}{c})^2)^3\big)  & \text{if $|x|\leq c$} \\[2pt]
		\frac{c^2}{6} &\text{otherwise} 
	\end{cases},
\end{equation}
where $c$ is a constant provided by the user.
It is typically set to the maximum residual error expected for inliers.

Based on the \ac{PDF} in (\ref{eq:c22}), the gradient and Hessian required for Newton optimization can be derived.
Similar to other methods \cite{Lepetit2005}, we first reformulate the Tukey norm as a re-weighted quadratic expression
\begin{equation}\label{eq:c25}
	\rho_\textrm{tuk}\big(\big\lVert\pmb{x}'_i - \pmb{x}_i(\pmb{\theta})\big\rVert_2\big) \approx w_i \big\lVert\pmb{x}'_i - \pmb{x}_i(\pmb{\theta})\big\rVert_2^2,
\end{equation}
with
\begin{equation}\label{eq:c26}
	w_i = \frac{\rho_\textrm{tuk}(r_i)}{r_i}, \quad r_i = \big\lVert\pmb{x}'_i - \pmb{x}_i\big\rVert_2^2.
\end{equation}
Weights $w_i$ are considered to be constant and are recalculated in each iteration of the optimization.
For their calculation, residuals $r_i$ are evaluated at $\pmb{\theta} = \pmb{0}$.
Finally, given the simplification, the gradient and Hessian can be derived as
\begin{align}\label{eq:c27}
	\pmb{g}_\textrm{t} =& \sum_{i=0}^{n} -\frac{w_i}{{\sigma_\textrm{t}}^2}
	\big(\pmb{x}_i - \pmb{x}'_i\big)^{\hspace{-1pt}\top}
	\frac{\partial \pmb{x}_i}{\partial {}_\textrm{C}\pmb{X}_i}
	\frac{\partial {}_\textrm{C}\pmb{X}_i}{\partial \pmb{\theta}}
	\bigg\vert_{\pmb{\theta}=\pmb{0}}\hspace{-1pt},\\\label{eq:c28}
	\pmb{H}_\textrm{t} \approx& \sum_{i=0}^{n} -\frac{w_i}{{\sigma_\textrm{t}}^2}
	\left(\frac{\partial \pmb{x}_i}{\partial {}_\textrm{C}\pmb{X}_i}
	\frac{\partial {}_\textrm{C}\pmb{X}_i}{\partial \pmb{\theta}}\right)^{\hspace{-3pt}\top}\hspace{-3pt}
	\left(\frac{\partial \pmb{x}_i}{\partial {}_\textrm{C}\pmb{X}_i}
	\frac{\partial {}_\textrm{C}\pmb{X}_i}{\partial \pmb{\theta}}\right)
	\hspace{-2pt}\bigg\vert_{\pmb{\theta}=\pmb{0}}\hspace{-1pt},
\end{align}
where second-order partial derivatives with respect to $\pmb{x}_i$ are neglected.
Also, starting from the definitions in (\ref{eq:c01}) to (\ref{eq:c04}), we are able to calculate the required partial derivatives
\begin{align}
	\frac{\partial \pmb{x}_i}{\partial {}_\textrm{C}\pmb{X}_i} &= \frac{1}{{}_\textrm{C}Z_i^2} \begin{bmatrix}
		{}_\textrm{C}Z_i f_x & 0 & -{}_\textrm{C}X_i f_x \\
		0 & {}_\textrm{C}Z_i f_y & -{}_\textrm{C}Y_i f_y
	\end{bmatrix},\\
	\frac{\partial {}_\textrm{C}\pmb{X}_i}{\partial \pmb{\theta}} &= 
	{}_\textrm{C}\pmb{R}_\textrm{M}
	\begin{bmatrix} -[_\textrm{M}\pmb{X}_i]_\times & \pmb{I}_3\end{bmatrix}.
\end{align}
In conclusion, given the derived gradient and Hessian, the developed modality can be integrated into our multi-modality tracker, allowing to consider texture information.

\subsection{Multi-region Tracking}\label{sec:c3}
In the following, we build on our method \textit{SRT3D} \cite{Stoiber2021} and its implementation in \textit{ICG}'s region modality \cite{Stoiber2022}.
While most region-based methods use dense information, \textit{SRT3D} features a sparse formulation that considers region information along so-called correspondence lines.
For an efficient representation of the object's geometry, a sparse viewpoint model is used.
It contains precomputed 3D contour points and normal vectors for multiple views all around the object.
During tracking, 3D points and normal vectors from the view closest to the current pose estimate are projected into the image to define the location and orientation of so-called correspondence lines.
For each line, a discrete probability distribution for the contour location is computed based on probabilities that pixels along the line belong to the foreground or background.
The pose is then optimized so that locations of projected 3D contour points jointly maximize the respective probabilities.
During tracking, the definition and maximization are iteratively repeated, using decreasing line scales.
An example of the approach is shown in Fig.~\ref{fig:in00}.
The formulation assumes that only one transition along each correspondence line exists.
To validate this assumption, the sparse viewpoint model contains the distances along the normal vector for which the foreground and background are not interrupted by each other.
If a distance is below a certain threshold, the correspondence line is rejected.

For our multi-region approach, we consider each region on the object's surface independently.
This means that only the respective object surface is regarded as foreground while everything else is defined as background.
The technique is very similar to conventional region-based tracking, which differentiates between the object and the background.
Consequently, we can directly adopt the formulation of \textit{ICG} \cite{Stoiber2022} and deploy one modality per region.
To represent the geometry of individual regions, we again use sparse viewpoint models.
However, since multiple connected geometries are considered, the generation process has to be adapted.

\begin{figure}[t]
	\centering
	\input{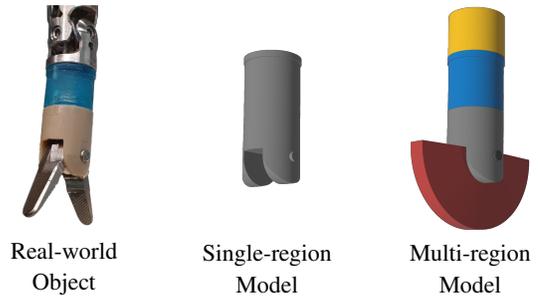}
	\caption{
		Region models of a robotic end effector.
		The image in the middle shows the 3D model that is used for single-region tracking.
		In the image on the right, the geometry is separated into blue and gray surface patches that are employed for multi-region tracking.
		In addition, a fixed yellow cylinder for the gimbal joint and a red shell geometry for the movable pliers are used to explicitly consider neighboring structures.
	}
	\label{fig:c34}
\end{figure}

\begin{figure}[t]
	\centering
\begingroup%
  \makeatletter%
  \providecommand\color[2][]{%
    \errmessage{(Inkscape) Color is used for the text in Inkscape, but the package 'color.sty' is not loaded}%
    \renewcommand\color[2][]{}%
  }%
  \providecommand\transparent[1]{%
    \errmessage{(Inkscape) Transparency is used (non-zero) for the text in Inkscape, but the package 'transparent.sty' is not loaded}%
    \renewcommand\transparent[1]{}%
  }%
  \providecommand\rotatebox[2]{#2}%
  \newcommand*\fsize{\dimexpr\f@size pt\relax}%
  \newcommand*\lineheight[1]{\fontsize{\fsize}{#1\fsize}\selectfont}%
  \ifx\svgwidth\undefined%
    \setlength{\unitlength}{127.55905512bp}%
    \ifx\svgscale\undefined%
      \relax%
    \else%
      \setlength{\unitlength}{\unitlength * \real{\svgscale}}%
    \fi%
  \else%
    \setlength{\unitlength}{\svgwidth}%
  \fi%
  \global\let\svgwidth\undefined%
  \global\let\svgscale\undefined%
  \makeatother%
  \begin{picture}(1,0.64444444)%
    \lineheight{1}%
    \setlength\tabcolsep{0pt}%
    \put(1.16872849,0.56957955){\makebox(0,0)[lt]{\lineheight{1.25}\smash{\begin{tabular}[t]{l}\small\end{tabular}}}}%
    \put(0,0){\includegraphics[width=\unitlength,page=1]{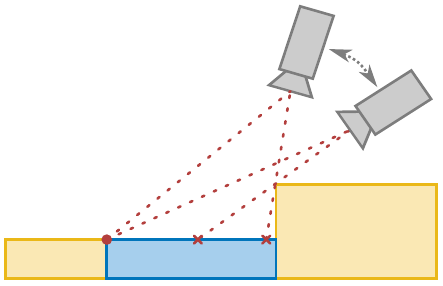}}%
  \end{picture}%
\endgroup%

	\caption{
		Contour validation for a blue region neighbored by two yellow regions.
		The left contour point, which is shown as a circle, is invariant to small camera changes and is therefore accepted.
		The point on the right, which is marked by a cross, emerges from an elevated edge of a neighboring geometry.
		It is highly view-dependent and is therefore rejected.
	}
	\label{fig:c30}
\end{figure}

In contrast to single-region tracking, sparse viewpoint models for multi-region tracking need to consider individual surface patches instead of the entire object.
An example of this is shown in Fig.~\ref{fig:c34}.
The geometry of surface patches is defined in separate meshes, which are created by the user.
During model generation, those meshes are rendered from multiple views all around the object to produce silhouette and depth images.
For a particular surface patch that models a region, the silhouette image shows a distinct pixel value.
This allows a simple extraction of the required contour.
For the extracted contour, it is important to validate points and reject view-dependent segments that are caused by shadows from neighboring geometries.
An example of this is shown in Fig.~\ref{fig:c30}.
For the validation, one simply compares the depth value of each contour point with that of neighboring pixels from other regions.
If the depth of those neighboring pixels is smaller than what can be explained by a typical surface gradient, the contour point is rejected.

\begin{figure}[t]
	\centering
	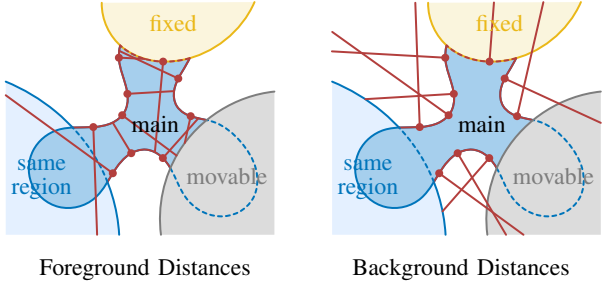
	\caption{
		Example for the validation of contour points and the computation of foreground and background distances.
		The valid contour is colored in red.
		Contour segments that are occluded by a \textit{movable} body or that neighbor a \textit{same-region} body are invalid.
		Segments on the transition to a \textit{fixed} body are only valid if they are invariant to small view changes.
		For the computation of foreground distances, only the \textit{main} and \textit{same-region} areas can be traversed.
		Background distances are stopped by any surface that considers the same region as the main geometry.
	}
	\label{fig:c31}
\end{figure}

In addition to \textit{fixed} neighboring regions, we also consider \textit{same-region} geometries that model color statistics similar to the main region and \textit{movable} external bodies.
Both cases are important for the tracking of robots.
For \textit{movable} bodies, one simply needs to create a shell mesh that covers the potentially occupied volume.
An example is shown in Fig.~\ref{fig:c34}, where a red mesh considers movable pliers.
The geometry is used during the validation to reject areas of the contour that could become occluded.
Also, in the case of \textit{same-region} surfaces, it does not make sense to deploy correspondence lines if regions with similar color statistics neighbor the main region.
Affected contour areas are thus also rejected.
Examples of both cases are shown in Fig.~\ref{fig:c31}.

While for the validated contour, points and normal vectors can be sampled as usual, it is also essential to correctly calculate foreground and background distances.
Background distances, which are oriented outwards, are limited either by the main region itself or by other \textit{same-region} areas.
Inwards-oriented foreground distances can only traverse the main region or \textit{same-region} bodies that are fixed.
Examples for both cases are again shown in Fig.~\ref{fig:c31}.
Finally, with the developed sparse viewpoint model, we are able to consider multiple regions while taking into account neighboring geometries.


\section{Implementation}\label{sec:i}
In the following, we provide implementation details and parameter values.
Our algorithm builds on \textit{ICG} and is publicly available\footnote{\url{https://github.com/DLR-RM/3DObjectTracking}}.
If not stated otherwise, default settings defined in the publication of \textit{ICG} \cite{Stoiber2022} are used.
For the texture modality, we integrate the \textit{BRISK} \cite{Leutenegger2011}, \textit{ORB} \cite{Rublee2011}, \textit{FREAK} \cite{Alahi2012}, \textit{SIFT} \cite{Lowe2004}, and \textit{DAISY} \cite{Tola2010} descriptors from \textit{OpenCV}.
\textit{FREAK} and \textit{DAISY} are used in combination with the highly-efficient \textit{ORB} detector.
All other algorithms provide their own detection method.
In general, we found that while \textit{SIFT} achieves a slightly higher score, \textit{ORB} provides the best trade-off between quality and efficiency.
A detailed comparison is given in Section~\ref{ssec:e2}.
For \textit{ORB}, a maximum of $300$ feature points are retrieved, and $3$ scale levels that differ by a factor of $1.2$ are used.
Also, \textit{SIFT} is run with $3$ octave layers, a contrast threshold of $0.04$, an edge threshold of $10$, and a sigma value of $0.7$.
For parameter values of other descriptors, please refer to the source code.

During matching, we use the Hamming distance for binary descriptors and the euclidean norm for \textit{SIFT} and \textit{DAISY}.
A match is valid if the distance ratio to the second-best match is smaller than $0.7$.
As input to the detector, we use a rectangular image crop that is based on the previous pose estimate, and that is cropped and scaled to a defined size of $200 \times 200\,\unit{px}$.
A frame is considered a keyframe if the orientational difference to the previous keyframe is bigger than $10^\circ$.
To check if a point from the keyframe is occluded, we use the same strategy as \textit{ICG} and reject points where the expected depth is significantly smaller than that of corresponding measurements.
In the evaluation, a Tukey norm constant of $c = 20\,\unit{px}$ is used.
For the user-defined standard deviation $\sigma_\textrm{t}$, which is given in units of pixels, we use $\{10, 10, 3\}$ for the \textit{YCB-Video} and $\{5, 1, 0.5\}$ for the \textit{OPT} dataset.
The given sets define parameters for each iteration, with the last value used for all remaining iterations.


\section{Evaluation}\label{sec:e}

In this section, we present a detailed evaluation on the \textit{YCB-Video} \cite{Xiang2018} and \textit{OPT} \cite{Wu2017} datasets, where we compare our approach to the current state of the art.
In addition, we conduct an ablation study that evaluates different feature descriptors and shows the importance of individual modalities.
Qualitative results for the \textit{YCB-Video} dataset and real-world applications are shown in a publicly available video.\footnote{\url{https://www.youtube.com/watch?v=NfNzxXupX54}}

\subsection{YCB-Video Dataset}\label{ssec:e0}
\begin{figure}[t]
	\centering
	\input{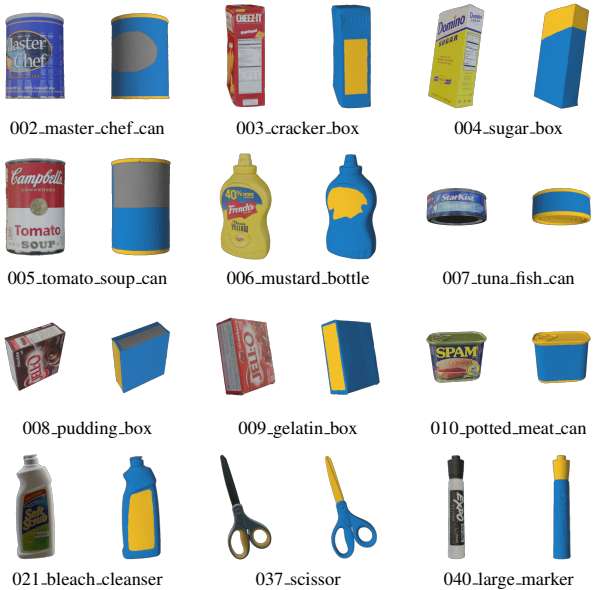}
	\caption{
		\textit{YCB} objects that are suitable for multi-region tracking.		
		The right image for each object illustrates the modeled regions in different colors.
	}\label{fig:e00}
\end{figure}

The \textit{YCB-Video} dataset \cite{Xiang2018} uses 21 \textit{YCB} objects and considers 12 sequences with 2949 frames in the evaluation.
Based on ground-truth poses, average distance and average shortest distance errors $e_\text{ADD}$ and $e_\text{ADD-S}$ can be calculated.
They were proposed by \cite{Hinterstoisser2013} and are defined as follows
\begin{align}\label{eq:e00}
	e_\text{ADD} &= \frac{1}{n}\sum_{i=1}^n\big\lVert \big({}_\text{M}\pmb{\widetilde{X}}_i -  {}_\text{M}\pmb{T}_{\text{M}_\text{GT}}\,{}_\text{M}\pmb{\widetilde{X}}_i\big)_{3\times 1}\big\rVert_2,\\\label{eq:e01}
	e_\text{ADD\textnormal{-}S} &= \frac{1}{n}\sum_{i=1}^n \min_{j\in[n]} \big\lVert \big({}_\text{M}\pmb{\widetilde{X}}_i -  {}_\text{M}\pmb{T}_{\text{M}_\text{GT}}\,{}_\text{M}\pmb{\widetilde{X}}_j\big)_{3\times 1}\big\rVert_2,
\end{align}
where ${}_\textrm{M}\pmb{T}_{\textrm{M}_{\textrm{GT}}}$ is the transformation from the ground-truth to the estimated model frame, $\pmb{X}_i$ is a 3D mesh vertex, $n$ is the number of vertices, and $()_{3\times 1}$ denotes the first three elements of a vector.
For the evaluation, ADD and ADD-S errors are used in an area under curve score.
Based on $m$ frames and a threshold of $e_\textrm{t} = 0.1\,\unit{m}$, it is calculated as
\begin{equation}\label{eq:e02}
	s = \frac{1}{m}\sum_{i=1}^m \max\Big(1 - \frac{e_{i}}{e_\textrm{t}}, 0\Big).
\end{equation}
The error $e_i$ is thereby either the ADD error $e_{\textrm{ADD}}$ or the ADD-S error $e_{\textrm{ADD-S}}$ for frame $i$.
Finally, for the evaluation of our multi-region approach, 12 suitable objects with distinctive regions were identified.
An overview of those objects and the modeled regions is shown in Fig.~\ref{fig:e00}.

Results from the evaluation are shown in Tab.~\ref{tab:e01}.
\begin{table*}
	\vspace{1ex}
	\caption{
		Results on the \textit{YCB-Video} dataset \cite{Xiang2018}, showing ADD and ADD-S area under curve scores in percent and framerates in hertz
	}\label{tab:e01}

\centering
\scriptsize
\begin{tabularx}{\textwidth}{X *{7}{| >{\centering\arraybackslash}p{0.84cm}@{\hspace{0.0cm}} >{\centering\arraybackslash}p{0.84cm}}}
\hline
\noalign{\smallskip}
\multirow{2}{1cm}{\textbf{Approach}} & 
\multicolumn{2}{c|}{\multirow{2}{1.64cm}{\centering  \!\!\! PoseCNN + ICP \!\!\! + DeepIM \cite{Xiang2018}}}&
\multicolumn{2}{c|}{\multirow{2}{1.64cm}{\centering DeepIM \cite{Li2018}}}&
\multicolumn{2}{c|}{\multirow{2}{1.64cm}{\centering se(3)-TrackNet \cite{Wen2020}}}&
\multicolumn{2}{c|}{\multirow{2}{1.64cm}{\centering PoseRBPF + SDF \cite{Deng2021}}}&
\multicolumn{2}{c|}{\multirow{2}{1.64cm}{\centering ICG \cite{Stoiber2022}}}&
\multicolumn{2}{c|}{\multirow{2}{1.64cm}{\centering ICG+ ORB (Ours)}}&
\multicolumn{2}{c}{\multirow{2}{1.64cm}{\centering ICG+ SIFT (Ours)}}\\
&\multicolumn{2}{c|}{}&
\multicolumn{2}{c|}{}&
\multicolumn{2}{c|}{}&
\multicolumn{2}{c|}{}&
\multicolumn{2}{c|}{}&
\multicolumn{2}{c|}{}&
\multicolumn{2}{c}{}\\
\noalign{\smallskip}
\hline
\noalign{\smallskip}
Initial Pose& \multicolumn{2}{c|}{--}  &\multicolumn{2}{c|}{Ground Truth} &\multicolumn{2}{c|}{Ground Truth} &\multicolumn{2}{c|}{PoseCNN} &\multicolumn{2}{c|}{Ground Truth} &\multicolumn{2}{c}{Ground Truth} &\multicolumn{2}{c}{Ground Truth}\\
Re-initialization& \multicolumn{2}{c|}{--}& \multicolumn{2}{c|}{Yes (290)}& \multicolumn{2}{c|}{No}& \multicolumn{2}{c|}{Yes (2)}&  \multicolumn{2}{c|}{No}& \multicolumn{2}{c|}{No}&  \multicolumn{2}{c}{No}\\
\noalign{\smallskip}
\hline
\noalign{\smallskip}
Objects & ADD & ADD-S & ADD & ADD-S & ADD & ADD-S & ADD & ADD-S & ADD & ADD-S & ADD & ADD-S & ADD & ADD-S \\
\noalign{\smallskip}
\hline
\noalign{\smallskip}
002\_master\_chef\_can  & 78.0 & 96.3 & 89.0 & 93.8 & 93.9 & 96.3 & 89.3 & 96.7 & 66.4 & 89.7 & \underline{94.0} & \underline{97.9} & \textbf{94.7} & \textbf{98.0}\\
003\_cracker\_box  & 91.4 & 95.3 & 88.5 & 93.0 & \underline{96.5} & 97.2 & 96.0 & 97.1 & 82.4 & 92.1 & 96.3 & \underline{98.2} & \textbf{96.8} & \textbf{98.4}\\
004\_sugar\_box  & \textbf{97.6} & 98.2 & 94.3 & 96.3 & \textbf{97.6} & 98.1 & 94.0 & 96.4 & 96.1 & \underline{98.4} & 96.0 & 98.3 & 96.6 & \textbf{98.5}\\
005\_tomato\_soup\_can  & 90.3 & 94.8 & 89.1 & 93.2 & \underline{95.0} & 97.2 & 87.2 & 95.2 & 73.2 & 97.3 & 94.8 & \textbf{98.1} & \textbf{95.2} & \textbf{98.1}\\
006\_mustard\_bottle  & \underline{97.1} & 98.0 & 92.0 & 95.1 & 95.8 & 97.4 & \textbf{98.3} & \underline{98.5} & 96.2 & 98.4 & 96.2 & 98.4 & \underline{97.1} & \textbf{98.7}\\
007\_tuna\_fish\_can  & 92.2 & \textbf{98.0} & 92.0 & 96.4 & 86.5 & 91.1 & 86.8 & 93.6 & 73.2 & 95.8 & \textbf{94.5} & \underline{97.2} & \underline{94.1} & 97.1\\
008\_pudding\_box  & \underline{83.5} & 90.6 & 80.1 & 88.3 & \textbf{97.9} & \textbf{98.4} & 60.9 & 87.1 & 73.8 & 88.9 & 80.6 & \underline{90.9} & 80.4 & 90.5\\
009\_gelatin\_box  & \underline{98.0} & 98.5 & 92.0 & 94.4 & 97.8 & 98.4 & \textbf{98.2} & 98.6 & 97.2 & \underline{98.8} & 96.9 & \underline{98.8} & 96.8 & \textbf{99.1}\\
010\_potted\_meat\_can  & 82.2 & 90.3 & 78.0 & 88.9 & 77.8 & 84.2 & 76.4 & 83.5 & 93.3 & 97.3 & \underline{94.8} & \underline{97.9} & \textbf{95.4} & \textbf{98.1}\\
011\_banana  & \underline{94.9} & 97.6 & 81.0 & 90.5 & \underline{94.9} & 97.2 & 92.8 & 97.7 & \textbf{95.6} & \textbf{98.4} & 94.1 & \underline{98.2} & 92.8 & \underline{98.2}\\
019\_pitcher\_base  & \underline{97.4} & 97.9 & 90.4 & 94.7 & 96.8 & 97.5 & \textbf{97.7} & 98.1 & 97.0 & \underline{98.8} & 97.0 & \underline{98.8} & 97.2 & \textbf{98.9}\\
021\_bleach\_cleanser  & 91.6 & 96.9 & 81.7 & 90.5 & \textbf{95.9} & \underline{97.2} & \textbf{95.9} & 97.0 & 92.6 & \textbf{97.5} & 90.1 & 96.5 & 91.6 & 97.0\\
024\_bowl  & 8.1 & 87.0 & 38.8 & 90.6 & 80.9 & 94.5 & 34.0 & 93.0 & 74.4 & \textbf{98.4} & \textbf{85.9} & \underline{97.9} & \underline{84.1} & \underline{97.9}\\
025\_mug  & 94.2 & 97.6 & 83.2 & 92.0 & 91.5 & 96.9 & 86.9 & 96.7 & \textbf{95.6} & \textbf{98.5} & 94.1 & 98.3 & \textbf{95.6} & \underline{98.4}\\
035\_power\_drill  & \underline{97.2} & 97.9 & 85.4 & 92.3 & 96.4 & 97.4 & \textbf{97.8} & 98.2 & 96.7 & \underline{98.5} & 96.5 & 98.4 & 96.8 & \textbf{98.7}\\
036\_wood\_block  & 81.1 & 91.5 & 44.3 & 75.4 & \textbf{95.2} & 96.7 & 37.8 & 93.6 & 93.5 & \underline{97.2} & \underline{94.2} & \textbf{97.5} & 91.7 & 96.7\\
037\_scissors  & 92.7 & 96.0 & 70.3 & 84.5 & \textbf{95.7} & \underline{97.5} & 72.7 & 85.5 & 93.5 & 97.3 & 93.9 & 97.4 & \underline{94.9} & \textbf{97.6}\\
040\_large\_marker  & 88.9 & \textbf{98.2} & 80.4 & 91.2 & \underline{92.2} & 96.0 & 89.2 & 97.3 & 88.5 & 97.8 & 88.2 & 97.3 & \textbf{94.0} & \underline{98.1}\\
051\_large\_clamp  & 54.2 & 77.9 & 73.9 & 84.1 & \textbf{94.7} & 96.9 & 90.1 & 95.5 & 91.8 & 96.9 & \underline{94.1} & \underline{97.7} & 93.8 & \textbf{97.8}\\
052\_extra\_large\_clamp  & 36.5 & 77.8 & 49.3 & 90.3 & \underline{91.7} & \underline{95.8} & 84.4 & 94.1 & 85.9 & 94.3 & 88.0 & 95.2 & \textbf{91.8} & \textbf{97.0}\\
061\_foam\_brick  & 48.2 & 97.6 & 91.6 & 95.5 & 93.7 & 96.7 & 96.1 & 98.3 & \textbf{96.2} & \underline{98.5} & \textbf{96.2} & \textbf{98.6} & 94.0 & 97.9\\
\noalign{\smallskip}
\hline
\noalign{\smallskip}
\textbf{All Frames}  & 80.7 & 94.0 & 82.3 & 91.9 & 93.0 & 95.7 & 87.5 & 95.2 & 86.4 & 96.5 & \underline{93.7} & \underline{97.7} & \textbf{94.3} & \textbf{97.9}\\
\noalign{\smallskip}
\hline
\noalign{\smallskip}
\textbf{FPS [Hz]} &\multicolumn{2}{c|}{0.1} &\multicolumn{2}{c|}{12.0} &\multicolumn{2}{c|}{90.9} &\multicolumn{2}{c|}{7.6} &\multicolumn{2}{c|}{\textbf{788.4}} &\multicolumn{2}{c|}{\underline{312.4}} &\multicolumn{2}{c}{111.7}\\
\noalign{\smallskip}
\hline
\end{tabularx}
	\vspace{-2.5ex}
\end{table*}
Our approach \textit{ICG+} is compared to the current state of the art in deep learning-based tracking \cite{Li2018, Wen2020, Deng2021}, as well as the previously best-performing conventional method \textit{ICG} \cite{Stoiber2022}.
The evaluation demonstrates that, both with \textit{ORB} and \textit{SIFT}, our method outperforms the competition by a significant margin.
Especially for the ADD metric, a huge improvement compared to the original \textit{ICG} approach without multi-region and texture can be observed.
Furthermore, \textit{ICG+} also improves on results from state-of-the-art object detection algorithms such as \textit{CosyPose} \cite{Labbe2020}, \textit{FFB6D} \cite{He2021}, and \textit{PR-GCN} \cite{Zhou2021}, which report average ADD-S scores of 89.8, 95.5, and 95.8 respectively.
To the best of our knowledge, \textit{ICG+} achieves the currently highest score among all methods.

In addition to tracking quality, we also compare average framerates.
For this, \textit{ICG+} was evaluated on a computer with an \textit{Intel Core i9-11900K} CPU and a \textit{NVIDIA RTX A5000} GPU.
Note that other methods were evaluated on different computers.
Nevertheless, the results demonstrate that, while the proposed extension of \textit{ICG} requires additional computational resources, \textit{ICG+} is still highly competitive.
Utilizing a single CPU core, the tracker achieves an average framerate of $312.4\,\unit{Hz}$ for the combination with \textit{ORB} and $111.7\,\unit{Hz}$ together with \textit{SIFT}.
Also, compared to deep learning-based methods, which heavily depend on parallel computation, in our approach, the GPU is only required for the occasional reconstruction of 3D points in keyframes.
As a consequence, it remains mostly idle.

\subsection{OPT Dataset}\label{ssec:e1}
The \textit{OPT} dataset \cite{Wu2017} uses $6$ objects and consists of $552$ real-world sequences that feature various trajectories, speeds, and lighting conditions.
In contrast to the \textit{YCB-Video} dataset, it is mostly used by conventional methods.
For the evaluation, the ADD area under curve score, which is based on Eqs.~(\ref{eq:e00}) and (\ref{eq:e02}), is used.
The threshold is thereby defined as $e_\textrm{t} = 0.2d$, where $d$ is the largest distance between model vertices.
Values are then scaled between $0$ and $20$ and are referred to as AUC scores.
Note that since most objects do not contain distinct regions, multi-region tracking is not considered.
Results of the evaluation are shown in Tab.~\ref{tab:e10}.
\begin{table}
	\caption{
		Results on the \textit{OPT} dataset \cite{Wu2017}, showing AUC scores
	}\label{tab:e10}

\centering
\scriptsize
\begin{tabularx}{\linewidth}{@{\hspace{0.15cm}} X@{\hspace{0.0cm}} *{6}{>{\centering\arraybackslash}p{0.88cm}@{\hspace{0.0cm}}} >{\centering\arraybackslash}p{0.88cm}@{\hspace{0.02cm}}}
\hline
\noalign{\smallskip}
\textbf{Approach}&Soda & Chest & Ironman & House & Bike & Jet &\textbf{Avg.}\\
\noalign{\smallskip}
\hline
\noalign{\smallskip}
PWP3D \cite{Prisacariu2012}&5.87&5.55&3.92&3.58&5.36&5.81&5.01\\
ORB-SLAM2 \cite{MurArtal2017}&13.44&15.53&11.20&17.28&10.41&9.93&12.97\\
Bugaev \cite{Bugaev2018}&14.85&14.97&14.71&14.48&12.55&17.17&14.79\\
Tjaden \cite{Tjaden2018}&8.86&11.76&11.99&10.15&11.90&13.22&11.31\\
Zhong \cite{Zhong2020}&9.01&12.24&11.21&13.61&12.83&15.44&12.39\\
Li \cite{Li2021}&9.00&14.92&13.44&13.60&12.85&10.64&12.41\\
Huang \cite{Huang2021} & 9.07 & 12.93 & 8.80 & 11.15 & 7.96 & 11.09 & 10.17 \\
SRT3D \cite{Stoiber2021}&15.64&16.30&17.41&16.36&13.02&15.64&15.73\\
ICG \cite{Stoiber2022}&15.32&15.85&17.86&17.92&16.36&15.90&16.54\\
ICG+ ORB (Ours)& \underline{16.51} & \underline{16.97} & \textbf{18.29} & \underline{18.59} & \textbf{17.13} & \textbf{17.91} & \textbf{17.57}\\
ICG+ SIFT (Ours)& \textbf{16.72} & \textbf{17.37} & \underline{17.93} & \textbf{18.64} & \underline{16.92} & \underline{17.82} & \textbf{17.57}\\
\noalign{\smallskip}
\hline
\end{tabularx}
\end{table}%
The evaluation includes state-of-the-art methods that use different sources of information such as region, edge, texture, and depth.
Results show that our approach achieves the currently highest score for all $6$ objects.

\subsection{Ablation Study}\label{ssec:e2}
For feature descriptors and detectors, many approaches exist.
We thus conduct a comparison of prominent methods and report results in Tab.~\ref{tab:e20}.
\begin{table}
	\caption{
		Comparison of different feature descriptors
	}\label{tab:e20}

\scriptsize
\begin{tabularx}{\linewidth}{X@{\hspace{0.1cm}} | >{\centering\arraybackslash}p{1.3cm} | *{2}{>{\centering\arraybackslash}p{1.1cm}@{\hspace{0.25cm}}}
		>{\centering\arraybackslash}p{1.1cm}}
\hline
\noalign{\smallskip}
\textbf{Dataset} & OPT \cite{Wu2017} &\multicolumn{3}{c}{YCB-Video \cite{Xiang2018}} \\
\noalign{\smallskip}
\hline
\noalign{\smallskip}
\textbf{Features} & AUC & ADD & ADD-S & FPS [Hz] \\
\noalign{\smallskip}
\hline
\noalign{\smallskip}
ICG+ BRISK \cite{Leutenegger2011} & 17.16 & \underline{94.2} & \underline{97.8} & 121.2\\
ICG+ ORB \cite{Rublee2011} & \textbf{17.57} & 93.7 & 97.7 & \underline{312.4}\\
ICG+ FREAK \cite{Alahi2012} & 15.84 & 91.7 & 96.9 & \textbf{400.5}\\
ICG+ SIFT \cite{Lowe2004} & \textbf{17.57} & \textbf{94.3} & \textbf{97.9} & 111.7\\
ICG+ DAISY \cite{Tola2010} & 16.78 & 93.6 & 97.6 & 204.5\\
\noalign{\smallskip}
\hline
\end{tabularx}
\end{table}%
The experiment demonstrates that while \textit{SIFT} \cite{Lowe2004} achieves the highest score, \textit{BRISK} \cite{Leutenegger2011} and \textit{ORB} \cite{Rublee2011} also deliver excellent results.
At the same time, \textit{FREAK} \cite{Alahi2012} achieves the best runtime.
However, tracking quality is significantly worse.
Given those results, we believe that \textit{ORB} provides the best trade-off between quality and efficiency.
Consequently, it is used as default.

In addition, we also compare the importance of individual modalities.
For this, results are provided in Tab.~\ref{tab:e02}.
\begin{table}
	\caption{
		Ablation study for individual modalities
	}\label{tab:e02}

\scriptsize
\begin{tabularx}{\linewidth}{X@{\hspace{0.1cm}} | >{\centering\arraybackslash}p{1.2cm} | *{2}{>{\centering\arraybackslash}p{1.0cm}@{\hspace{0.25cm}}}
>{\centering\arraybackslash}p{1.0cm}}
\hline
\noalign{\smallskip}
\textbf{Dataset} & OPT \cite{Wu2017} &\multicolumn{3}{c}{YCB-Video \cite{Xiang2018}} \\
\noalign{\smallskip}
\hline
\noalign{\smallskip}
\textbf{Experiment} & AUC & ADD & ADD-S & FPS [Hz] \\
\noalign{\smallskip}
\hline
\noalign{\smallskip}
ICG+ ORB & 17.57 & 93.7 & 97.7 & 312.4\\
W/o Multi-region & - & 92.3 & 96.9 & 354.8\\
W/o Texture & 15.05 & 90.2 & 96.6 & 742.2\\
W/o Multi-region, Texture & 15.05 & 86.5 & 96.3 & 1035.7\\
W/o Depth & 17.40 & 47.5 & 60.9 & 426.1\\
W/o Region & 15.90 & 22.4 & 48.6 & 444.3\\
\noalign{\smallskip}
\hline
\end{tabularx}
\end{table}
Experiments show that, while multi-region and texture nicely complement each other, scores on the \textit{YCB-Video} dataset remain relatively high if either is disabled.
The main reason is that both help to overcome ambiguities in the object's geometry.
Each technique thereby has its own advantages.
For example, while the texture modality achieves better tracking results, the multi-region approach is robust to motion blur, extremely fast, and does not require a GPU during tracking.
Finally, the importance of the depth and region modality are also analyzed.
The obtained results demonstrate that, while  multi-modality tracking seems less important for the \textit{OPT} dataset, using all available information is essential for the more challenging sequences of the \textit{YCB-Video} dataset.


\section{Conclusion}\label{sec:d}

In this work, we extended the geometry-based algorithm \textit{ICG} to incorporate information from visual appearance using two different modalities.
The developed texture modality uses keypoint features and the object's geometry to estimate pose changes between frames.
Thanks to this relative formulation, no texture is required for the model, which improves usability.
Furthermore, the proposed multi-region approach allows the incorporation of additional region information within the object's contour.
Using multiple regions also leads to more coherent color statistics.
This further improves tracking results.
In addition, it is possible to model external geometries, which is especially important for the tracking of bodies in kinematic structures such as robots.
Finally, \textit{ICG+} is highly modular and allows a flexible combination of depth, texture, and multi-region information from multiple cameras.

Evaluations on the \textit{YCB-Video} and \textit{OPT} datasets show that the resulting algorithm significantly improves the state of the art and achieves the highest scores.
This is especially interesting since it demonstrates that deep learning-based methods do not yet surpass classical techniques, even with textured 3D models, significant computational resources, and vast training data.
Finally, our method is also highly efficient and runs at more than $300\,\unit{Hz}$.
Thanks to those properties, we are confident that \textit{ICG+} is able to support a wide variety of applications both in augmented reality and robotics.

\bibliographystyle{IEEEtran}
\bibliography{IEEEabrv, literature}

\end{document}